\documentclass[letterpaper]{article} % DO NOT CHANGE THIS
\usepackage{aaai25}  % DO NOT CHANGE THIS
\usepackage{times}  % DO NOT CHANGE THIS
\usepackage{helvet}  % DO NOT CHANGE THIS
\usepackage{courier}  % DO NOT CHANGE THIS
\usepackage[hyphens]{url}  % DO NOT CHANGE THIS
\usepackage{graphicx} % DO NOT CHANGE THIS
\usepackage{natbib}  % DO NOT CHANGE THIS AND DO NOT ADD ANY OPTIONS TO IT
\usepackage{caption} % DO NOT CHANGE THIS AND DO NOT ADD ANY OPTIONS TO IT
\usepackage{algorithm}
\usepackage{algorithmic}

\usepackage{newfloat}
\usepackage{listings}
\DeclareCaptionStyle{ruled}{labelfont=normalfont,labelsep=colon,strut=off} % DO NOT CHANGE THIS
\floatstyle{ruled}
\newfloat{listing}{tb}{lst}{}
\floatname{listing}{Listing}
\usepackage{amsthm}
\usepackage{amsmath}
\usepackage{amssymb}
\usepackage{amsfonts}
\usepackage{booktabs}
\usepackage{multirow}
\usepackage{enumitem}
\title{RUNA: Object-level Out-of-Distribution Detection via Regional
Uncertainty Alignment of Multimodal Representations}
\author{
    Bin Zhang\equalcontrib \textsuperscript{\rm 1,\rm 2},
    Jinggang Chen\equalcontrib \textsuperscript{\rm 1},
    Xiaoyang Qu\footnote{Corresponding authors}\textsuperscript{\rm 2},
    Guokuan Li\textsuperscript{\rm 1},\\
    Kai Lu\footnotemark[\value{footnote}]\textsuperscript{\rm 1},
    Jiguang Wan\textsuperscript{\rm 1},
    Jing Xiao\textsuperscript{\rm 2},
    Jianzong Wang\textsuperscript{\rm 2}
}
\affiliations{
    \textsuperscript{\rm 1}Wuhan National Laboratory for Optoelectronics, Huazhong University of Science and Technology, Wuhan, China\\
    \textsuperscript{\rm 2}Ping An Technology (Shenzhen) Co., Ltd, Shenzhen, China\\
    \{binz2398, chen.jinggang98, quxiaoy\}@gmail.com, liguokuan@hust.edu.cn,\\ \{kailu, jgwan\}@hust.edu.cn, xiaojing661@pingan.com, jzwang@188.com
}

\usepackage{bibentry}
\begin{document}

\maketitle

\begin{abstract}
Enabling object detectors to recognize out-of-distribution (OOD) objects is vital for building reliable systems. A primary obstacle stems from the fact that models frequently do not receive supervisory signals from unfamiliar data, leading to overly confident predictions regarding OOD objects. Despite previous progress that estimates OOD uncertainty based on the detection model and in-distribution (ID) samples, we explore using pre-trained vision-language representations for object-level OOD detection. We first discuss the limitations of applying image-level CLIP-based OOD detection methods to object-level scenarios. Building upon these insights, we propose RUNA, a novel framework that leverages a dual encoder architecture to capture rich contextual information and employs a regional uncertainty alignment mechanism to distinguish ID from OOD objects effectively. We introduce a few-shot fine-tuning approach that aligns region-level semantic representations to further improve the model's capability to discriminate between similar objects. Our experiments show that RUNA substantially surpasses state-of-the-art methods in object-level OOD detection, particularly in challenging scenarios with diverse and complex object instances.
\end{abstract}

% Uncomment the following to link to your code, datasets, an extended version or similar.
%
% \begin{links}
%     \link{Code}{https://aaai.org/example/code}
%     \link{Datasets}{https://aaai.org/example/datasets}
%     \link{Extended version}{https://aaai.org/example/extended-version}
% \end{links}

\section{Introduction}

Identifying out-of-distribution (OOD) objects is vital for object detectors to safely deploy in an open-world environment. Most current models work in a closed-world environment, matching objects to the pre-defined in-distribution (ID) labels. Nevertheless, when deployed in an open-world environment, they acknowledge the possibility of encountering objects from unknown categories, which should not be naively assigned to any ID labels. It poses a risk to the security of object detection models. In high-stakes applications like autonomous driving, failure to detect OOD objects can lead to severe accidents \cite{automot_OOD}. We can mitigate this risk if unknown objects are detected and the human driver is alerted to take control.

\begin{figure}[t]
\centering
\includegraphics[width=0.95\columnwidth]{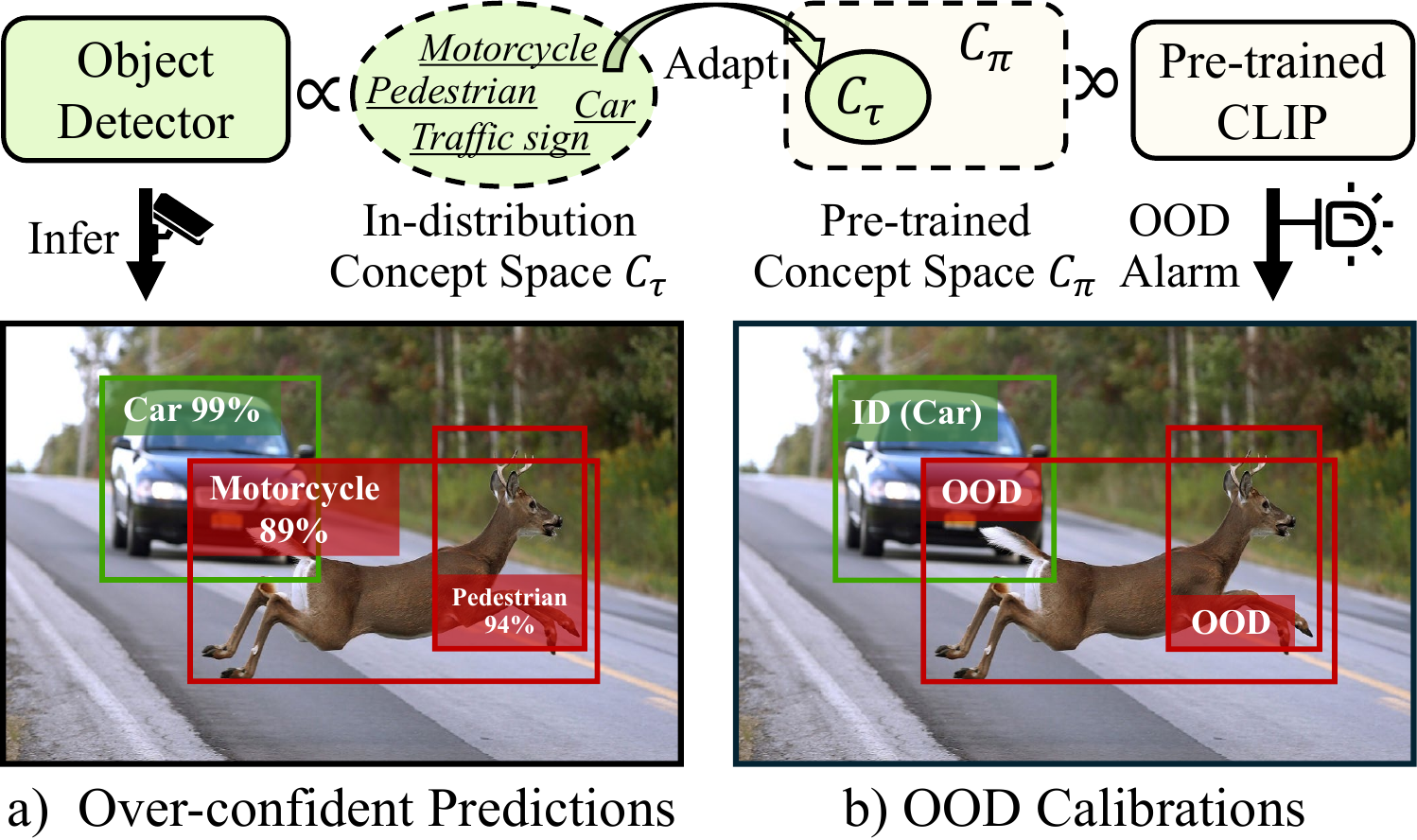}
\caption{Object detectors in the open world tend to make erroneous decisions when facing unknown objects, threatening machine learning system security. To mitigate this, we adapt knowledge-rich vision-language representations into the ID concept space for object-level OOD detection.}
\label{fig::illustration}
\end{figure}

The susceptibility to OOD inputs stems from insufficient knowledge about unknowns during training. It results in neural networks tending to abnormally generate over-confident predictions when semantic shifts occur within the samples \cite{OOD_overconfidence_hein2019relu}, as shown in Figure \ref{fig::illustration}. In object-level OOD detection, a line of prior works \cite{Siren, VOS, video_OOD, wu2023tib, wu2024unsupervised} design additional modules to be integrated into the training process of detectors, leveraging the model's inherent uncertainty. Meanwhile, some estimation-based methods directly learning from training data \cite{ma_distance, tack2020csi, sun2022out_knn} are also introduced into this domain as alternative solutions. 

In recent years, advancements in contrastive multimodal pre-training methods, including ALIGN \cite{ALIGN}, CLIP \cite{CLIP}, BLIP \cite{li2022blip} and InternVL \cite{chen2024internvl}, have provided a novel perspective for detecting out-of-distributions. With extensive prior knowledge, vision-language (VL) representations can transferably detect distributional shifts in downstream image-level classification tasks given the ID textual class labels \cite{CLIPOOD, MCM}. This observation prompts us to delve further: if this pre-trained alignment capability can be adapted to measure the regional uncertainty for individual objects, we would effortlessly unlock a safety assistant for deployed detectors, replacing previous limited enhancement methods and potentially boosting the performance of object-level OOD detection. However, applying these pre-trained models to the object-level OOD detection task presents substantial challenges. Unlike image-level classification, where the entire image is considered, object detection focuses on localized regions. This localization process can lead to a loss of contextual information, making it difficult to accurately assess an object's anomaly.

Moreover, the effectiveness of pre-trained models is often affected by the variety and quality of the dataset of the object detector. Datasets like BDD-100K\cite{yu2020bdd100k}, which contain a wide range of object sizes, lighting conditions, and occlusions, can pose challenges for models trained on more generic datasets. For instance, an object detector trained on BDD-100K might misclassify small, occluded objects from other datasets as ID vehicles. This situation underscores the necessity for domain-specific adaptation to enhance the performance of these models in applied settings.

This study proposes a novel framework, RUNA, to address the abovementioned limitations for object-level OOD detection. RUNA leverages a dual encoder architecture to provide rich contextual information and employs a \textbf{R}egional \textbf{UN}certainty \textbf{A}lignment strategy to effectively calculate uncertainty scores for object regions, enabling accurate classification as ID or OOD. We employ few-shot fine-tuning to bridge the performance gap between generic and domain-specific datasets. The diverse and challenging nature of datasets like BDD-100K, characterized by varying lighting conditions, occlusions, and object appearances, necessitates tailored model adaptation. Moreover, the pre-trained VL models, primarily trained on scene-centric images, exhibit limited alignment capability with the unique characteristics of domain-specific datasets. Fine-tuning allows the model to acquire domain-specific characteristics, enhancing its capability to detect OOD objects efficiently.

Our contributions are as follows:
\begin{itemize}[nosep]
    \item We propose RUNA, a novel object-level OOD detection framework with a dual encoder architecture that captures global and local features for accurate regional uncertainty estimation.
    \item We develop a few-shot fine-tuning approach to efficiently align region-level ID semantics, substantially enhancing the model's capacity to differentiate between ID and OOD objects.
    \item Our approach remarkably improves object-level OOD detection performance compared to previous methods.
    % It achieves an average decrease of 16\% FPR95 on VOC and 10.9\% on BDD-100K.
\end{itemize}

\section{Preliminaries}

\textbf{Object-level OOD Detection.} Object-level OOD detection aims to identify unknown objects that fall outside the recognized categories and may be misidentified by the model. Object-level OOD detection is more applicable to real-world machine learning systems than image-level OOD detection. However, it also presents more significant challenges, as it requires careful consideration of each object's uncertainties at a fine-grained level.

We define the ID space as $\mathcal{X}$, with the associated label space given by $\mathcal{Y}_{\text{in}}=\{ y_1, y_2, y_3, ... y_K\}$. In object-level OOD detection literature\cite{Siren, video_OOD, VOS, wu2023tib, wu2024unsupervised}, regional objects ${ \hat{x}_b }$ from an OOD sample $x_{\text{out}}$ are considered to experience a semantic shift compared to ID objects, meaning their label space $\mathcal{Y}_{\text{out}}$ does not overlap with $\mathcal{Y}_{\text{in}}$.

For an unknown image $x$, the object detector $f_{\theta}$ predicts results as $D_x = \{b_i, y^{p}_{i}\}_{i=1}^m$. Here, $b_i \in \mathbb{R}^{4}$ represents the bounding box, and $y^{p}_{i} \in \mathcal{Y}_{\text{in}}$ is its ID semantic label. OOD detection is structured as a binary classification task with uncertainty estimation $\sigma(\cdot)$, distinguishing between ID and OOD objects. Given a bounding box $b$, the goal is to predict the uncertainty $\sigma(x, b)$:
\begin{equation}
G(\hat{x}_b, \mathcal{Y}_{\text{in}}) = \begin{cases}
\text{in}, &\text{if}\quad \mathbb{E}[\sigma(x, b) \mid \mathcal{Y}_{\text{in}}] \leq \gamma\\
\text{out}, &\text{if}\quad \mathbb{E}[\sigma(x, b) \mid \mathcal{Y}_{\text{in}}] > \gamma\\
\end{cases}
\end{equation}
% \begin{equation}
% G = \begin{cases}
% \text{ID}, &\text{if}\quad U \leq \gamma\\
% \text{OOD}, &\text{if}\quad U > \gamma\\
% \end{cases}
% \end{equation}
where $\gamma$ is the threshold chosen such that a high fraction of ID data (e.g., 95\%) falls below it.

\begin{figure}[t]
\centering
\includegraphics[width=0.95\columnwidth]{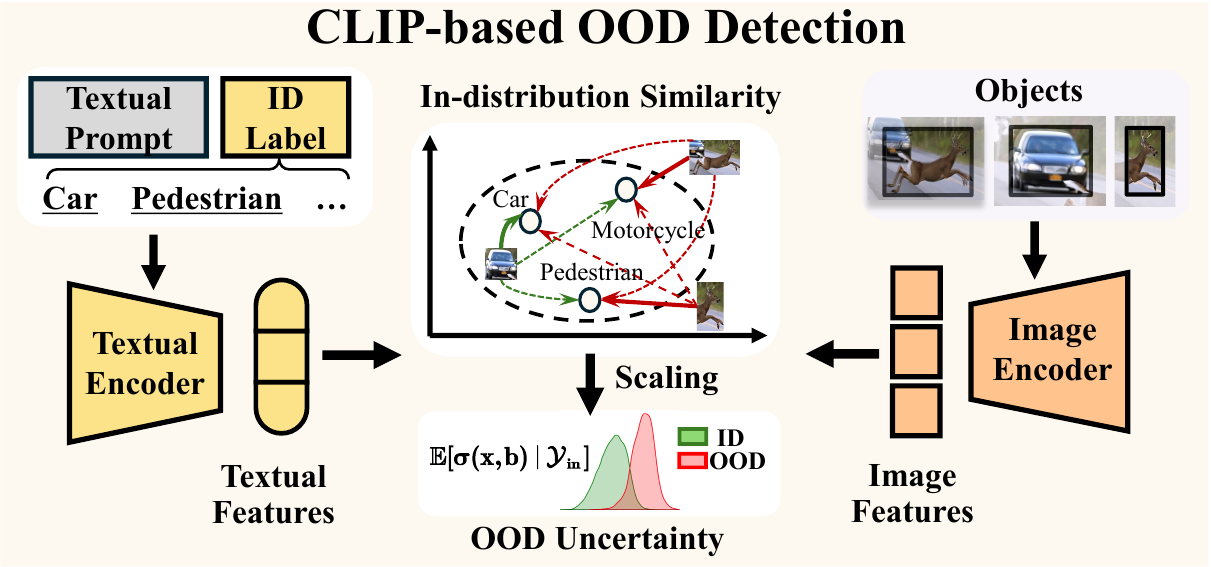}
\caption{Framework of CLIP-based OOD Detection. Green arrows represent ID samples, while red arrows denote OOD samples. The solid line highlights the maximum similarity, and the dotted lines indicate other similarity measures.}
\label{fig::zero-shot_framework}
\end{figure}

\textbf{Zero-shot CLIP-based OOD Detection.} CLIP has excelled in zero-shot OOD detection tasks by utilizing extensive training data and large-scale models.

We outline the approach for performing zero-shot OOD detection utilizing existing CLIP-based zero-shot methodologies Maximum Concept Matching (MCM)\cite{MCM}. As depicted in Figure \ref{fig::zero-shot_framework}, the CLIP model employs an image encoder to extract features from images. While CLIP lacks an explicit classifier, we can use ID classes to create text inputs (e.g., "a photo of a dog"). The text encoder processes these text inputs to produce class-specific features, which act similarly to a classifier. Let $\mathcal{T}_{\text{in}}$ represent the collection of test prompts that include $K$ class labels. To illustrate, \citeauthor{MCM} calculates the score with the softmax score of the similarity between image features and textual features:
\begin{equation}
    S_{\text{MCM}}(x)=\max _{t_i \in \mathcal{T}_{\text {in }}} \frac{e^{\text{Sim}(\mathcal{I}(x),\mathcal{T}(t_i))  / \tau}}{\sum_{t_c \in \mathcal{T}_{\text {in }}} e^{\text{Sim}(\mathcal{I}(x),\mathcal{T}(t_c)) / \tau}}
    \label{equa:MCM}
\end{equation}
where $x$ denotes the input image, $t_i$ and $t_c$ denotes the text to match. $\text{Sim}(\mathcal{I}(x),\mathcal{T}(t_i))$ denotes the similarity between image feature $\mathcal{I}(x)$ and textual feature $\mathcal{T}(t_i)$. If the MCM for a given image is below a predefined threshold, it is classified as ID; otherwise, it is considered OOD. 

\begin{figure*}[t]
\centering
\includegraphics[width=0.99\textwidth]{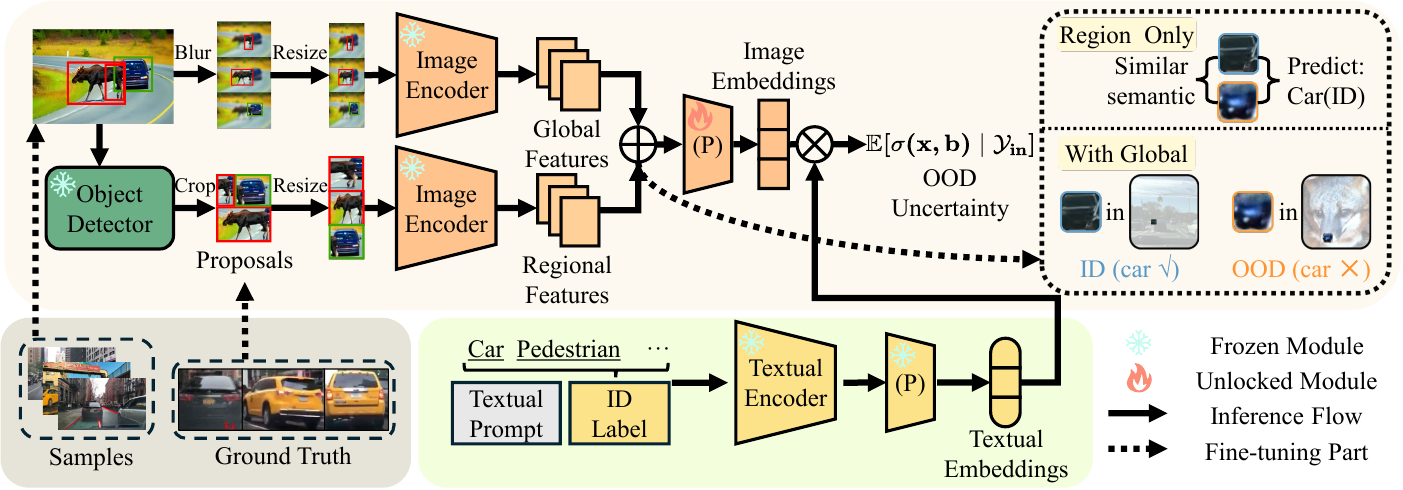} 
\caption{\textbf{Overview of RUNA Framework.} Our novel dual-encoder architecture computes regional object uncertainty by extracting global and regional image features and aligning them with text features. During fine-tuning, the image encoder handling regional images remain frozen, while only its projection layer (P) participate in the fine-tuning. The upper right dashed box highlights the importance of our feature fusion strategy: when encountering objects with similar semantics, limited local features can lead to incorrect decisions. By incorporating global features, the model can make more informed judgments.}

% We aim to achieve a more sensitive alignment effect on ID regional images through effective fine-tuning. The object detector, the text encoder and the image encoder processing global images remain frozen all the time. For the image encoder handling regional images, the vision transformer layers remain frozen, while only the projection layer (P) participates in the fine-tuning process. The red line in upper right dashed box highlights two process: the right focuses on improving the inter-class discriminative ability of ID samples, while the left further strengthens the model's ID sensitivity to discriminate between ID and OOD samples assisted by global features.
\label{fig::overview}
\end{figure*}

% \section{RUNA: Fine-Tuning Framework with Regional Uncertainty Alignment}
\section{Method}

\textbf{Overview.} As in Figure \ref{fig::overview}, our framework, RUNA, utilizes a dual Image Encoder structure ${\mathcal{I}^{(g)}, \mathcal{I}^{(r)}}$ to collaboratively process visual information, where $\mathcal{I}^{(g)}$ captures global features from the entire image, and $\mathcal{I}^{(r)}$ focuses on regional features by processing specific objects or areas of interest. The outputs are subsequently fused to produce the final image embeddings. We employ a metric based on the maximum similarity score to quantify uncertainty in object detection. Notably, we exclusively fine-tune the projection layer of the visual encoder, denoted as $\mathcal{I}_{P}$, to tailor the pre-trained model for our particular task.

Given an image $x$ and its corresponding region of interest (bounding box) $\hat{x}_b$, our framework first extracts global features from the entire image using $\mathcal{I}^{(g)}(x)$. Here, bounding box blurring is applied to keep contextual information while emphasizing the objects of interest, enhancing the extracted features' relevance. Concurrently, $\mathcal{I}^{(r)}(\hat{x}_b)$ processes the region of interest by cropping the specific area, enabling a more focused and detailed representation of the object.

To effectively integrate the semantic information extracted from the regional image with the global context, we propose a novel fusion strategy, which can be expressed as:
\begin{equation}
\mathcal{I}_{\text{t}}(x, \hat{x}_{b}) = \mathcal{I}_P(\lambda \cdot \mathcal{I}^{(r)}(\hat{x}_{b}) \oplus (1-\lambda) \cdot \mathcal{I}^{(g)}(x))
\end{equation}
% \begin{equation}
% \mathcal{I}_{\text{t}}(x, \hat{x}_{b}) = \mathcal{I}_P( \mathcal{I}^{(r)}(\hat{x}_{b}) \oplus  \mathcal{I}^{(g)}(x))
% \end{equation}
where $\oplus$ denotes element-wise addition, and $\lambda$ is a weighting coefficient that regulates the influence of each encoder. The resulting $\mathcal{I}_{\text{t}}(x, \hat{x}_{b})$ represents the final image embedding, which synthesizes both regional and global features.

As illustrated in the top-right dashed box on the left in Figure \ref{fig::overview}, when encountering objects with similar semantics, relying solely on limited local features may cause the model to misinterpret subtle differences, leading to incorrect decisions. This is especially problematic when visually similar objects share overlapping attributes, making it challenging to distinguish between them using only localized cues. By incorporating global features, the model gains a broader context, enabling it better to grasp the overall scene structure and relational information. This holistic view allows for more informed and accurate judgments, as the model can integrate detailed local patterns and the larger contextual backdrop, leading to more robust OOD detection.

% \textbf{In-Distribution Semantic Space Construction.}
As illustrated in Figure \ref{fig::overview}, our approach functions as a post-hoc corrective technique that does not interfere with the training process of the object detection model. For a regional object $\hat{x}_b$ identified by the detector’s prediction ${x, b}$, we transform the distributional uncertainty $\sigma(x, b)$ of $\hat{x}_b$ into a distance measure relative to the ID semantic space. Given the ID space as $\mathcal{P} \in \mathbb{R}^{K}$, the uncertainty of a predicted region $\hat{x}_b$ is represented as its deviation from this known distribution $\mathcal{P}$:
\begin{equation}
\mathbb{E}[\sigma(x, b) \mid \mathcal{Y}_{\text{in}}] = \mathbb{E}_{\mathcal{P}}[\mathcal{H}(\hat{x}_b, \mathcal{P})]
\end{equation}
where $\mathcal{H}$ denotes the selected distance measurement function. In this framework, the uncertainty of an unknown object is converted into a distance-based estimation relative to the ID space $\mathcal{P}$. In line with prior research\cite{MCM}, we construct the $K$ dimensions of $\mathcal{P}$ using the pre-trained CLIP text encoder $\mathcal{T}(\cdot)$.

\textbf{Regional Uncertainty Alignment.} Our goal is to quantify the discrepancy between the input region image $\hat{x}_b$ and the entire ID semantic space $\mathcal{P}$. This discrepancy is accomplished by assessing the similarity between the image features and predefined concept vectors corresponding to ID labels, with this similarity serving as a distance measure.

Initially, we consider that the semantic similarities between $\mathcal{I}_{\text{t}}(x, \hat{x}_{b})$ and all ID concept vectors contribute to assessing its distance from the ID space $\mathcal{P}$ which means $\mathcal{T}(t_i)$. A straightforward method is to sum these similarities (Direct Sum) to reflect the degree of deviation of $\{x,\hat{x}_b\}$:
\begin{equation}
\mathbb{E}  [ \sigma(x, b) \mid \mathcal{Y}_{\text{in}} ] =  -\sum_{i=1}^{K} \text{Sim}(\mathcal{I}_{\text{t}}(x, \hat{x}_{b}),\mathcal{T}(t_i))
\end{equation}

The negative sign indicates lower semantic similarity with ID concept vectors for OOD objects corresponds to a greater distance from $\mathcal{P}$, implying higher uncertainty.

However, during our evaluation, we observed that Direct Sum fails to effectively distinguish between ID and OOD objects in the VOC dataset. We attribute this issue to the limited variance in the cosine similarities outputted by CLIP, which do not exhibit significant differences between matching and non-matching situations. This results in the absolute values of the summed similarities, overshadowing the impact of actual differences. 

\begin{figure}[t]
    \centering
    \includegraphics[width=0.98\columnwidth]{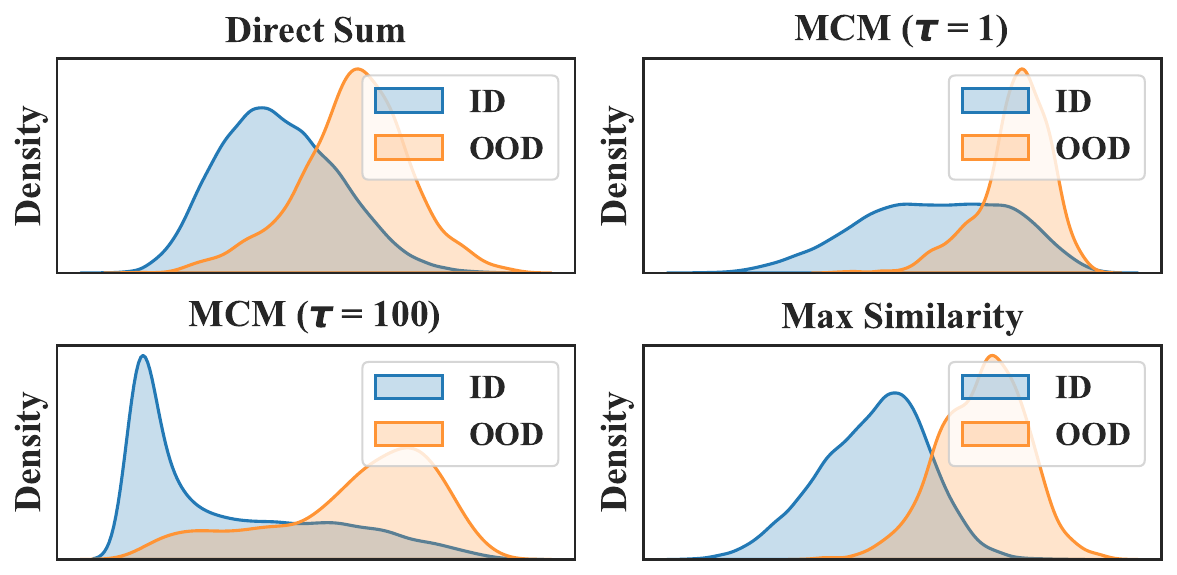}
    \caption{Distribution of uncertainty scores for Direct Sum, MCM($\tau$ = 1), MCM($\tau$ = 100) and Max Similarity.}
    \label{fig:id_ood_distribution}
\end{figure}

To tackle this issue, we adapt the MCM in Eq.(\ref{equa:MCM}) by substituting $\mathcal{I}(x)$ with $\mathcal{I}_t(x,\hat{x}_b)$, thereby amplifying the differences between the similarities. However, MCM does not significantly improve the differentiation between ID and OOD objects, as shown in Figure \ref{fig:id_ood_distribution}. Interestingly, we notice that as the scaling factor of MCM increases, its performance improves. This phenomenon prompts us to investigate further. We find that as the differences between values expand, the influence of larger values on the scores also increases. When the factor approaches its limit, the score is predominantly influenced by the maximum similarity value. Consequently, we define the uncertainty estimation metric as follows:
\begin{equation}
\mathbb{E} [ \sigma(x, b) \mid \mathcal{Y}_{\text{in}} ] = - \max_{1 \leq i \leq K} \text{Sim}(\mathcal{I}_{\text{t}}(x, \hat{x}_{b}),\mathcal{T}(t_i))
\end{equation}

% \begin{equation}
% U = - \max_{1 \leq i \leq K} \text{Sim}(\mathcal{I}_{\text{t}}(x, \hat{x}_{b}),\mathcal{T}(t_i))
% \end{equation}
where $\text{Sim}(\mathcal{I}_{\text{t}}(x, \hat{x}_{b}),\mathcal{T}(t_i))$ denotes similarity between the image feature $\mathcal{I}_{\text{t}}(x, \hat{x}_{b})$ and the concept vector $\mathcal{T}(t_i)$ for each label $i$ and $K$ denotes the number of labels.

\textbf{Few-shot Fine-tuning.} Although the zero-shot method lays a solid groundwork for OOD detection, it lacks the nuanced understanding of ID data needed to distinguish between ID and OOD samples precisely. We introduce a fine-tuning strategy that utilizes few-shot learning for cost-effective and rapid adaptation to fill this gap. By randomly selecting a small set of images, our approach infuses the model with region-specific details, unlocking a more profound comprehension of ID characteristics.

For a given image $x$, we treat all ground truth bounding boxes $\{\hat{x}_b^i\}_{i=1}^m$ as potential fine-tuning candidates. From these, $N$ shots of $K$ kinds of objects are randomly drawn, denoted as $\{(\hat{x}^{i}_{b}, y_i)\}_{i=1}^{NK}$, where each $\hat{x}_b^i$ corresponds to a regional patch and $y_i$ indicates its associated label. This process selectively exposes the model to key ID regions, enhancing its ability to align with the fine-grained semantic features vital for robust ID/OOD discrimination.

Given the pre-trained image encoder’s capacity for intense feature extraction, our fine-tuning selectively focuses on refining the projection layer after the image encoder handles regional images. We aim to align visual embeddings with their corresponding label embeddings closely. To this end, we employ a contrastive loss that sharpens the model’s intra-ID discriminative power:
\begin{equation}
    \mathcal{L}_{\text{ID}}=-\sum_{\hat{x}_b \in \mathcal{B}  } \log \frac{\exp ( \text{Sim}(\mathcal{I}(\hat{x}_{b}),\mathcal{T}(t_i)) / \tau )}{\sum_{j=1}^{K} \exp ( \text{Sim}(\mathcal{I}(\hat{x}_{b}),\mathcal{T}(t_j)) / \tau )}
    \label{equ:loss}
\end{equation}
where $\tau$ is a temperature to scale cosine similarities. This contrastive loss formulation hones the model’s precision within the ID space and optimizes its ability to differentiate between subtle category variations, driving a more context-aware and semantically aligned fine-tuning process.
\begin{table*}[t]
\centering
\begin{tabular}{@{}clcclcc@{}}
\toprule
\multirow{2}{*}{\shortstack{In-distribution\\datasets}}          & \multirow{2}{*}{Detection Method}    & \multicolumn{2}{c}{OpenImages} & & \multicolumn{2}{c}{MSCOCO}  \\ \cmidrule(lr){3-4} \cmidrule(l){6-7} 
 &                                                & FPR95 $\downarrow$ & AUROC $\uparrow$ & & FPR95 $\downarrow$ & AUROC $\uparrow$  \\ \midrule
\multirow{12}{*}{\shortstack{\textbf{Berkeley}\\\textbf{DeepDrive-100k}}} & MSP \cite{MaxConfidenceScore} & 79.04          & 77.38         & & 80.94        & 75.87         \\
 & ODIN \cite{ODIN}                        & 58.92              & 76.61            & & 62.85              & 74.40                     \\
 & Mahalanobis \cite{ma_distance}          & 60.16              & 86.88            & & 57.66              & 84.92                     \\
 & Energy score \cite{liu2020energy}       & 54.97              & 79.60            & & 60.06              & 77.48                     \\
 & Gram matrices \cite{Gram}               & 77.55              & 59.38            & & 60.93              & 74.93                     \\
 & Generalized ODIN \cite{baseline_G-ODIN} & 50.17              & 87.18            & & 57.27              & 85.22                     \\
 & CSI \cite{Baseline_CSI_2020}            & 37.06              & 87.99            & & 47.10              & 84.09                     \\
 & SIREN \cite{Siren}                      & 37.19              & 87.87            & & 39.54              & 88.37                     \\
 & VOS \cite{VOS}                          & 35.61              & 88.46            & & 44.13              & 86.92                     \\
 & $\text{MCM}^{*}$ \cite{MCM}                         & 45.37              & 88.46            & & 53.79              & 86.92                     \\
 & SAFE \cite{wilson2023safe}                      & 16.04              & 94.64      &      & 32.56              & 88.96                     \\
 & TIB \cite{wu2023tib}                       & 24.00              & 92.54            & & 36.85              & 88.47                     \\ 
 & PCA-based  \cite{wu2024unsupervised}                     & 35.05              & 88.92            & & 45.72              & 85.14                     \\
 % & RONIN  \cite{nguyen2024zero}                     & 30.00              & 91.60            & 30.16              & 92.77            &         \\
  & \textbf{RUNA (Ours)}     & \textbf{9.95}     & \textbf{96.76}   & & \textbf{16.85}     & \textbf{93.92}            \\ \midrule
\multirow{12}{*}{\textbf{PASCAL-VOC}}                          & MSP \cite{MaxConfidenceScore} & 73.13          & 81.91         & & 70.99        & 83.45         \\
 & ODIN \cite{ODIN}                        & 63.14              & 82.59            & & 59.82              & 82.20                     \\
 & Mahalanobis \cite{ma_distance}          & 96.27              & 57.42            & & 96.46              & 59.25                     \\
 & Energy score \cite{liu2020energy}       & 58.69              & 82.98            & & 56.89              & 83.69                     \\
 & Gram matrices \cite{Gram}               & 67.42              & 77.62            & & 62.75              & 79.88                     \\
 & Generalized ODIN \cite{baseline_G-ODIN} & 70.28              & 79.23            & & 59.57              & 83.12                     \\
 & CSI \cite{Baseline_CSI_2020}            & 57.41              & 82.95            & & 59.91              & 81.83                     \\
 & SIREN \cite{Siren}                      & 49.12              & 87.21            & & 54.23              & 86.89                     \\
 & VOS \cite{VOS}                          & 50.79              & 85.42            & & 47.29              & 88.35            \\
 & $\text{MCM}^{*}$ \cite{MCM}                          & 48.73              & 80.16            & & 50.43              & 78.22                     \\
 & SAFE \cite{wilson2023safe}                      & \textbf{20.36}              & 92.28      &      & 47.40              & 80.30                     \\
 & TIB \cite{wu2023tib}                       & 47.19              & 88.09            & & 41.55              & 90.36                     \\ 
 & PCA-based  \cite{wu2024unsupervised}                     & 50.56              & 85.71            & & 44.54              & 89.40                     \\
 % & RONIN  \cite{nguyen2024zero}                     & 18.91              & 92.69            & 28.04              & 91.46            &         \\
 & \textbf{RUNA (Ours)}                           & 26.07     & \textbf{93.63}   & & \textbf{30.67}     & \textbf{92.48}                     \\ \bottomrule
\end{tabular}
\caption{\textbf{Main results.} $\uparrow$ denotes that higher values are considered superior, while $\downarrow$ signifies that lower values are desirable. All results are presented as percentages. \textbf{Bold} numbers represent superior results, and the second-best performance is marked with an \underline{underline}. $*$ means adapted with our dual-encoder architecture.}
\label{tab:main_tab}
\end{table*}

\section{Experiments}

% In this section, we first validate the effectiveness of our methods on object detection models. Then we conduct the ablation studies for further analysis.

\subsection{Experimental Settings}

% \textbf{Datasets and metrics.} For ID datasets, we use PASCAL-VOC \cite{VOC} and BDD-100K \cite{yu2020bdd100k}. We choose the detection model Faster R-CNN with ResNet50 \cite{resnet} backbone, which is pre-trained on these two ID datasets. Then we evaluate them on two  OOD datasets, which contain a subset of images from MS-COCO \cite{mscoco} and OpenImages \cite{openimages}, ensuring that labels associated with the objects within them do not intersect with the labels present in the ID datasets. Following the literature, we adopt three key metrics: 1) \textbf{FPR95}: The False Positive Rate (FPR) of OOD objects, when the True Positive Rate (TPR) of ID samples is maintained at 95\%. It can be understood as the proportion of OOD objects that are misclassified when the selected threshold correctly identifies most of the ID objects; 2) \textbf{AUROC}: The Area Under The Receiver Operating Characteristic curve (AUROC) is defined by the area under the ROC curve with TPR on the y-axis and FPR on the x-axis; higher is better. An AUROC score of 50\% indicates a method that is as effective as random guessing; 3) \textbf{mAP}: Our RUNA OOD detector is a post-hoc addition to a pre-trained object detection network and does not affect the on-task performance of the base model under the mean average precision (mAP) metric, as such, we do not report mAP as in VOS\cite{VOS}.

\textbf{Datasets and metrics.} We use PASCAL-VOC\cite{VOC} and BDD-100K\cite{yu2020bdd100k} as ID datasets and evaluate on two OOD datasets sourced from MS-COCO\cite{mscoco} and OpenImages\cite{openimages}, ensuring no label overlap with ID datasets. The object detection model is pre-trained on the ID datasets. We evaluate using three metrics: 1) \textbf{FPR95}: False Positive Rate at 95\% True Positive Rate for ID samples, indicating the proportion of misclassified OOD objects; 2) \textbf{AUROC}: Area Under the ROC curve, where higher values indicate superior performance; 3) \textbf{mAP}: We do not report mAP as the object detection model is not affected by the integreted RUNA OOD detector.

% \textbf{Models and baselines.} For the frozen detection model, we utilize the Detectron2 platform \cite{wu2019detectron2}, which is widely adopted in practical applications. Our experiments employ CLIP (VIT-B/16)\cite{CLIP} as the pre-trained vison-language model. And in the main results, we compare our CLIP-based methods with other widely used OOD detection methods, including Maximum Softmax Probability \cite{MaxConfidenceScore}, ODIN \cite{ODIN}, energy score \cite{liu2020energy}, Mahalanobis distance \cite{ma_distance}, Generalized ODIN \cite{baseline_G-ODIN}, CSI \cite{Baseline_CSI_2020} and Gram matrices \cite{Gram}. Moreover, we also compare with several object-level OOD detection methods (SIREN\cite{Siren}, VOS\cite{VOS}, TIB\cite{wu2023tib} and PCA-based method\cite{wu2024unsupervised}).
% , which require retraining the detection network. However, our CLIP-based approaches are under the category of post-hoc methods, ensuring higher detection efficiency without affecting the regular training of the detection model.

\textbf{Models and Baselines.} We utilize the Detectron2 platform \cite{wu2019detectron2} with Faster R-CNN\cite{ren2015faster_roi} (using ResNet50\cite{resnet} as the backbone) as the frozen object detection model. We adopt CLIP (VIT-B/16) \cite{CLIP} for the vision-language model. Our CLIP-based approaches are evaluated against several widely adopted image-level approaches, including MSP \cite{MaxConfidenceScore}, ODIN \cite{ODIN}, Mahalanobis \cite{ma_distance}, Generalized ODIN \cite{baseline_G-ODIN}, CSI \cite{Baseline_CSI_2020}, Gram matrices \cite{Gram}, Energy score \cite{liu2020energy}, and CLIP-based MCM\cite{MCM}. Additionally, we compare with object-level OOD detection methods such as VOS \cite{VOS}, SIREN \cite{Siren}, SAFE \cite{wilson2023safe}, TIB \cite{wu2023tib}, and PCA-based method \cite{wu2024unsupervised}.

\textbf{Implementation details.} We exclusively fine-tune the projection layer of the regional image encoder individually, while the global image encoder is off-the-shelf. For few-shot learning, we perform fine-tuning using 10-shot samples. For ID discriminative fine-tuning, we employ a batch size of 256 and use the AdamW optimizer, conducting fine-tuning over 100 epochs with a base learning rate of $5\times10^{-6}$. We use "a photo of a \{label\}" for the textual prompt. We set the dual encoder's fusion coefficient $\lambda$ to 0.5 and the blur radius R of Gaussian Blur to 1.

\subsection{Main Results}

\label{sec::main_results}

As illustrated in Table \ref{tab:main_tab}, our proposed CLIP-based approach, RUNA, demonstrates advantages over previous methods. Notably, on the autonomous driving dataset BDD-100K, our fine-tuning approach significantly enhances the detection of OOD objects. In tests on the OOD dataset OpenImages, RUNA achieves an FPR95 of \textbf{9.95\%}, marking a \textbf{6.09\%} reduction compared to the previously best-performing method, SAFE. On the OOD dataset MSCOCO, our method achieved an FPR95 of \textbf{16.85\%}, improving by \textbf{15.71\%} compared to SAFE. When VOC serves as the ID dataset and OpenImages as the OOD dataset, SAFE achieves a lower FPR95 compared to our method. However, on the OOD dataset MSCOCO, our method outperforms SAFE with an FPR95 of \textbf{30.67\%}, achieving an improvement of \textbf{16.73\%} over SAFE and \textbf{10.88\%} over SOTA method TIB. Our approach provides detector-agnostic performance without requiring manual feature selection, making it more flexible and broadly applicable across various detection scenarios. In contrast, previous methods explicitly designed for object-level OOD detection require retraining the target detection model, which may affect its original detection performance.
% When VOC serves as the ID dataset and OpenImages as the OOD dataset, RUNA performs the best, achieving an FPR95 of \textbf{26.07\%}, which is an improvement of \textbf{21.12\%} compared to TIB. On the OOD dataset MSCOCO, our method achieved an FPR95 of \textbf{30.67\%}, improving by \textbf{10.88\%} compared to TIB. It is also worth noting that previous methods designed explicitly for object-level OOD detection, such as VOS, SIREN, and TIB, necessitate retraining the target detection model, which could potentially impact the performance of the original model during detection tasks. 

\begin{table}[htbp]
\centering
\begin{tabular}{@{}cccccc@{}}
\toprule
\multirow{2}{*}{ID}  & \multicolumn{3}{c}{Method}                                                        & \multicolumn{2}{c}{OpenImages/MSCOCO} \\ \cmidrule(l){2-6} 
                     & RE                        & GE                        & FT                        & FPR95 $\downarrow$ & AUROC $\uparrow$ \\ \midrule
\multirow{5}{*}{\begin{tabular}[c]{@{}c@{}}BDD-\\ 100k\end{tabular}} &
  \checkmark &
  - &
  - &
  58.09/62.17 &
  70.94/69.10 \\
                     & -                         & \checkmark & -                         & 31.78/39.24        & 89.21/84.19      \\
                     & \checkmark & \checkmark & -                         & 29.42/36.96        & 89.90/85.37      \\
                     & \checkmark & -                         & \checkmark & 15.71/22.14        & 93.34/92.26      \\
 &
  \checkmark &
  \checkmark &
  \checkmark &
  \textbf{9.95/16.85} &
  \textbf{96.76/93.92} \\ \midrule
\multirow{5}{*}{VOC} & \checkmark & -                         & -                         & 42.75/55.26        & 91.19/84.65      \\
                     & -                         & \checkmark & -                         & 39.22/52.49        & 91.45/87.49      \\
                     & \checkmark & \checkmark & -                         & 38.97/51.37        & 91.78/88.92      \\
                     & \checkmark & -                         & \checkmark & 30.76/34.31        & 92.01/90.17      \\
 &
  \checkmark &
  \checkmark &
  \checkmark &
  \textbf{26.07/30.67} &
  \textbf{93.63/92.48} \\ \bottomrule
\end{tabular}
\caption{Ablation study on our regional encoder (RE), global encoder (GE) and few-shot fine-tune (FT). }
\label{tab:ablation_study_dual_ft}
\end{table}

\begin{figure*}[htbp]
    \centering
    \includegraphics[width=0.95\textwidth]{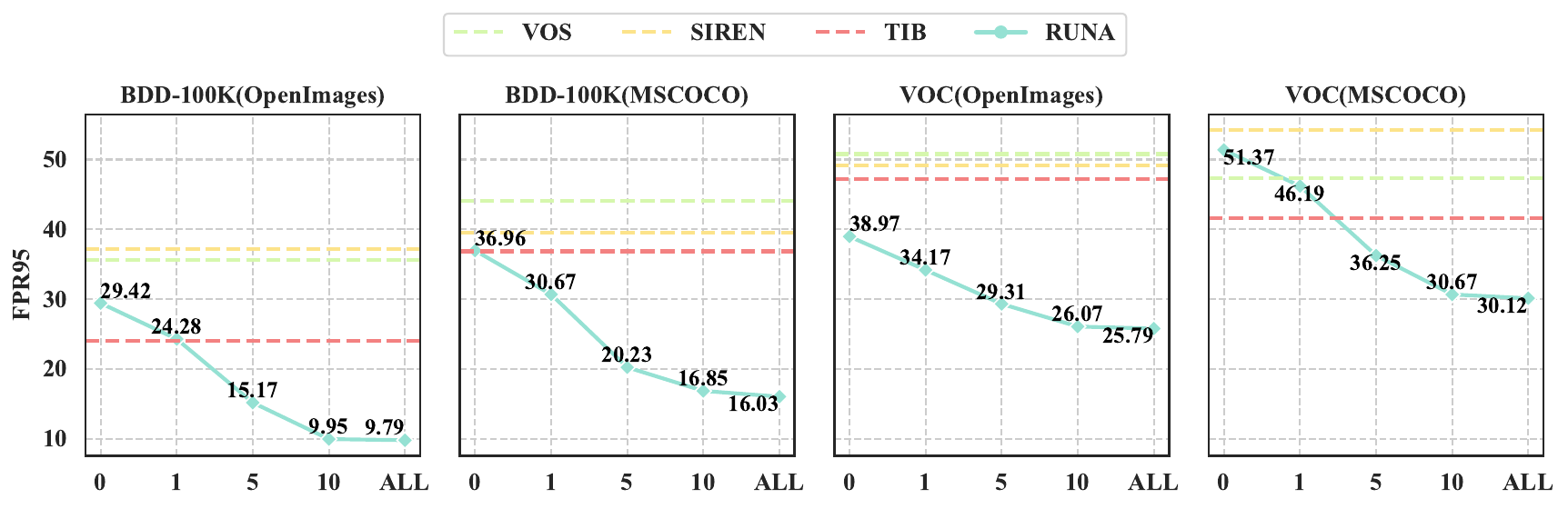}
    \caption{Ablation study on the number of fine-tuning samples (shots). This study examines how varying the number of shots affects detection performance, showing the trade-offs between data efficiency and detection quality.}
    \label{fig:ablation_sample_number}
\end{figure*}

\begin{table*}[htbp]
\centering
\begin{tabular}{@{}ccclcclcccc@{}}
\toprule
\multirow{2}{*}{Backbone} & \multicolumn{2}{c}{BDD-100k} &  & \multicolumn{2}{c}{VOC} &  & \multicolumn{3}{c}{Run time(ms)} & \multirow{2}{*}{Params} \\ \cmidrule(lr){2-3} \cmidrule(lr){5-6} \cmidrule(lr){8-10}
              & FPR95 $\downarrow$ & AUROC $\uparrow$ &  & FPR95 $\downarrow$ & AUROC $\uparrow$ &  & 1-image & 5-image & 10-image &       \\ \midrule
VIT-B/32      & 59.47/64.11      & 68.02/70.13    &  & 43.59/54.37      & 89.19/80.23    &  & 62.9    & 90.7    & 129.1    & 0.15B \\
VIT-B/16      & 58.09/62.17      & 70.94/69.10    &  & 42.75/55.26      & 91.19/84.65    &  & 83.8    & 220.1   & 410.4    & 0.15B \\
VIT-L/14      &  57.24/61.01     & 71.89/71.24    &  & 42.38/53.49      & 91.56/86.67    &  & 309.1   & 863.7   & 1559.5   & 0.43B \\
VIT-L/14(336) & 56.99/60.12      & 72.80/73.01    &  & 42.01/53.35      & 91.98/87.13    &  & 533.3   & 2086.5  & 4673.9   & 0.43B \\ \bottomrule
\end{tabular}
\caption{Ablation on the effect of different CLIP configurations on performance and run time. We evaluate the effects upon zero-shot regional encoder only method. The results are presented for two OOD datasets, with OpenImages followed by MSCOCO.}
\label{tab:ablation_study_clip_model}
\end{table*}

Furthermore, our fine-tuning framework showcases a remarkable improvement over the zero-shot approach. For instance, on VOC, RUNA shows a substantial FPR95 improvement of \textbf{12.90\%} on OpenImages and \textbf{20.70\%} on MSCOCO. Specifically, on BDD-100K, RUNA shows a substantial FPR95 improvement of \textbf{19.47\%} on OpenImages and \textbf{20.11\%} on MSCOCO.

\subsection{Ablation Study}

\label{sec::ablation_study}

\textbf{Ablation on the dual encoder and ID fine-tuning}. We examine the impact of the dual encoder and ID fine-tuning components within our object-level OOD detection framework. The fine-tuning strategy is designed to improve the discriminative capability of the pre-trained model with respect to ID semantics. At the same time, the dual encoder focuses on strengthening sensitivity to ID objects via enhanced feature representation. The evaluation results are presented in Table \ref{tab:ablation_study_dual_ft}. We compare the performance of models with and without the dual encoder to assess its contribution to feature extraction. Additionally, we assess the impact of fine-tuning on ID data, analyzing its role in adapting the model to the unique features of the ID space and improving OOD detection accuracy. The results demonstrate that the dual encoder significantly enhances the model's capability to distinguish between ID and OOD samples. ID fine-tuning further refines this capability, leading to more robust OOD detection.

\textbf{Ablation on the number of fine-tuning samples (shots).} We examine the effect of changing the number of samples used for model fine-tuning. The evaluation results are presented in Figure \ref{fig:ablation_sample_number}. The zero-shot method depends entirely on the pre-trained model, providing a competitive baseline performance but struggling with object-level OOD detection due to the lack of ID supervision. Introducing 1-shot learning shows immediate improvements, leveraging a single example to better align the model with the target task. With 5-shot learning, the model demonstrates significant gains, as many examples facilitate a more comprehensive understanding of the data distribution. Finally, 10-shot learning further enhances performance, capturing even more nuances and variations within the data. This study illustrates the clear benefits of incorporating a few labeled examples, with each incremental increase in sample size resulting in notable improvements in the model's accuracy and robustness. However, we also perform fine-tuning on the entire dataset, which yields only minimal improvements while incurring significantly more computational costs. This indicates that while few-shot learning provides substantial benefits, full dataset fine-tuning offers diminishing returns compared to the increased computational demands.

\textbf{Ablation on different backbones of the visual encoder.} We assess the influence of various ViT backbones on the performance of the CLIP model, as illustrated in Table \ref{tab:ablation_study_clip_model}. The analysis reveals that while the more extensive backbones, ViT-L/14 and ViT-L/14-336, provide slight improvements in FPR95 and AUROC, they substantially increase runtime and parameter count. Specifically, the ViT-L/14 and ViT-L/14-336 backbones, despite their slightly better performance, significantly increase computational demands. On the other hand, ViT-B/16 offers a good balance between efficiency and performance, demonstrating that the additional computational cost of larger models does not proportionally enhance performance. Consequently, we choose ViT-B/16 as the backbone for its optimal trade-off between performance gains and resource efficiency. This choice ensures that our framework remains computationally feasible while delivering high accuracy in OOD detection.

\section{Related Work}

\subsection{OOD Detection for Classification}

OOD detection distinguishes the unknown inputs that deviate from ID training data during the testing phase. The employment of the maximum softmax probability (MSP) \cite{MaxConfidenceScore} serves as a common baseline approach; however, it can yield excessively high values for OOD inputs \cite{OOD_overconfidence_hein2019relu}. Various enhancements have been suggested, such as ODIN \cite{ODIN}, Mahalanobis \cite{ma_distance}, Energy score \cite{liu2020energy}, Gram matrices \cite{Gram}, CSI \cite{Baseline_CSI_2020}, GODIN \cite{baseline_G-ODIN}, etc. While traditional OOD detection methods \cite{OE_dhamija2018reducing, OE_GANleetraining, OE_hendrycksdeep, OE_li2020background,chen2024gaia, Gen_kingma2018glow, Gen_schirrmeister2020understanding} largely stemmed from image-level classification tasks, some unique challenges posed by object detection require specialized approaches.

% Several improvements such as ODIN \cite{ODIN}, Mahalanobis \cite{ma_distance}, energy score \cite{liu2020energy}, Gram matrices score \cite{Gram}, have been proposed.

% While traditional OOD detection methods largely stemmed from image-level classification tasks, some unique challenges posed by object detection require specialized approaches.

%  Other branches such as outlier exposure methods regularization using natural images \cite{OE_dhamija2018reducing, OE_hendrycksdeep, OE_li2020background} or images synthesized by GANs \cite{OE_GANleetraining}, and generative models directly estimate the ID density \cite{Gen_kingma2018glow, Gen_schirrmeister2020understanding, Gen_van2016conditional}, which makes them natural alternatives for OOD detection. 

\subsection{Object-level OOD Detection}

OOD detection has garnered increasing attention in object detection to ensure the robustness of visual systems. Mainstream approaches \cite{Siren, VOS} primarily focus on model regularization to achieve optimal representations. In contrast, the SAFE method enhances OOD detection by selecting sensitivity-aware features from the object detector and inputting them into an auxiliary MLP network \cite{wilson2023safe}. Recent advancements include the Two-Stream Information Bottleneck method \cite{wu2023tib}, which utilizes dual information streams to identify unfamiliar objects without relying on auxiliary data, and the PCA-Driven Dynamic Prototype Enhancement technique, which dynamically refines prototypes for improved OOD discrimination using Principal Component Analysis \cite{wu2024unsupervised}. While many conventional methods depend on model uncertainty for OOD detection, we leverage a pre-trained vision-language model. The enhanced alignment knowledge from this model enables us to overcome the cognitive limitations of the original detection framework, resulting in improved OOD detection performance.

% OOD detection for object detection is a rising topic with very few existing works. 

\subsection{OOD Detection with Vision-language Models}

Recent vision-language models, such as CLIP \cite{CLIP}, have significantly advanced computer vision by aligning images and text in a shared feature space using a self-supervised contrastive objective. In OOD detection, \citeauthor{CLIPOOD} proposed ZOC, integrating a transformer-based decoder with CLIP’s image encoder \cite{CLIP}, tackling the challenge of sourcing OOD candidate labels—an issue our method bypasses. Building on this, \citeauthor{MCM} introduced a CLIP-based OOD detection approach using MCM, while \cite{miyai2023zero_multiobject} explored zero-shot ID detection to determine whether all objects in an image are ID.
Unlike image-level studies, our work harnesses vision-language models for object-level OOD detection. The core challenge is to localize and identify region-level out-of-distribution objects within images. This task becomes particularly complex when most pre-trained models are designed for image-level representations. In contrast to previous studies that heavily rely on textual prompting, we concentrate on visual prompting to effectively guide CLIP's attention to the target of interest. By integrating few-shot learning, we greatly enhance the efficiency of the fine-tuning process. This approach allows us to achieve optimal performance with a minimal amount of data, thereby improving the model's flexibility and robustness across various contexts and reassuring the reader of the reliability of our approach.
% Given the real-world implications of accurate object detection in open environments, our research in this direction holds considerable promise and value.

% \section{Future Work}

 % This involves refining the visual prompts to better focus on critical features and context, potentially improving the model's accuracy and robustness in complex scenarios.  This will require optimizing the model for efficient inference, reducing computational overhead, and ensuring robust performance in diverse, real-world conditions. By addressing these areas, we hope to expand the applicability of our approach and bring advanced OOD detection capabilities to practical, on-the-edge applications.

\section{Conclusion}

This paper focuses on detecting OOD objects using pre-trained vision-language representations. Our investigation begins with evaluating the effectiveness of CLIP-like representations in identifying object-level OOD instances and proposes the zero-shot baseline. Additionally, we propose RUNA, a novel Regional UNcertainty Alignment strategy, which significantly enhances detection performance by adapting vision-language models to ID semantic space and guiding vision-language models to be more sensitive to ID concepts. Overall, experimental results demonstrated substantial improvements over previous methods, emphasizing the effectiveness and promise of our proposed approaches. In the future, we intend to investigate more refined visual prompting techniques to enhance the model's capacity to capture subtle details and variations within the data. Furthermore, we aim to explore the deployment of our model to edge services, enabling real-time object detection and OOD detection in resource-constrained environments.

\section{Acknowledgement}

This research was funded by the Key Research and Development Program of Guangdong Province (grant No. 2021B0101400003) and the National Key Research and Development Program of China under (grant No.2023YFB4502701). This work was done while Bin Zhang was interning at Ping An Technology.

\bibliography{main}

\end{document}